\documentclass[letterpaper, 10 pt, conference]{ieeeconf}  

\IEEEoverridecommandlockouts                              

\usepackage{algorithm}
\usepackage{microtype}
\usepackage{graphicx}  
\usepackage{booktabs}  
\usepackage[]{algpseudocode}
\usepackage{amsfonts}
\usepackage{amsmath}
\usepackage{color}
\usepackage{soul}
\usepackage[authormarkuptext=name,addedmarkup=bf,authormarkupposition=left]{changes}
\usepackage{xspace}
\usepackage{hyperref}
\usepackage{cleveref}
\usepackage{dsfont}
\usepackage{bm}
\usepackage{bbm}
\usepackage{nicefrac}  
\usepackage[scaled=.8]{beramono}  

\usepackage{amsthm}
\usepackage{amsfonts}
\usepackage{amssymb}
\usepackage{array}
\usepackage{mathtools}
\usepackage{caption}
\usepackage{subcaption}
\usepackage{fontawesome}
\usepackage{thmtools}
\usepackage{thm-restate}
\usepackage{xspace}
\usepackage{lipsum}
\usepackage{pifont}
\usepackage{makecell}
\usepackage{tablefootnote}
\usepackage[tight-spacing=true]{siunitx}
\usepackage{multirow}
\usepackage{afterpage}
\usepackage{tabularx} 
\usepackage{url}
\usepackage{tikz}
\usepackage{gensymb} 
\usepackage{mathtools}
\usepackage{balance}

\DeclarePairedDelimiterX{\infdivx}[2]{(}{)}{%
  #1\;\delimsize\|\;#2%
}

\newcommand\mydots{\hbox to 1em{.\hss.\hss.}}
\newcommand{\suchthat}{\;\ifnum\currentgrouptype=16 \middle\fi|\;}

\Crefname{figure}{Figure}{Figures}
\crefname{figure}{Figure}{Figures}
\crefname{table}{Table}{Tables}
\crefformat{table}{Table~#1}
\crefformat{algorithm}{Algorithm~#1}
\crefformat{equation}{Eq.~(#1)}
\crefformat{section}{Section~#1}

\newcommand{\paperTitle}{
    Policy Distillation and Value Matching in\\Multiagent Reinforcement Learning
}

\newcommand\sdots{\hbox to 1em{.\hss.\hss.}} 

\usetikzlibrary{decorations.pathreplacing,calc}
\newcommand{\tikzmark}[1]{\tikz[overlay,remember picture] \node (#1) {};}
\newcommand*{\AddNote}[4]{%
    \begin{tikzpicture}[overlay, remember picture]
        \draw [decoration={brace,mirror,amplitude=0.5em},decorate,thick,black]
            ($(#3)!(#1.north)!($(#3)-(0,1)$)$) --  
            ($(#3)!(#2.south)!($(#3)-(0,1)$)$)
                node[align=center,text width=2.5cm,pos=0.5,anchor=east,xshift=0.65cm] {#4};
    \end{tikzpicture}
}%

\newif\ifcomments
\commentsfalse 

\ifcomments
	\newcommand{\swXX}[1]{\color{olive}SW: (#1)\color{black}\xspace}  
	\newcommand{\dXX}[1]{\color{red}DK: (#1)\color{black}\xspace}  
	\newcommand{\soXX}[1]{\color{cyan}SO: (#1)\color{black}\xspace}  
	\newcommand{\XX}[1]{\color{orange}JH: (#1)\color{black}\xspace}  
\else
    \newcommand{\swXX}[1]{}  
    \newcommand{\dXX}[1]{}  
	\newcommand{\soXX}[1]{}  
	\newcommand{\XX}[1]{}  
\fi 

\title{\LARGE \bf \paperTitle}

\author{Samir Wadhwania, Dong-Ki Kim, Shayegan Omidshafiei$^\dagger$\thanks{$^\dagger$ Work was completed while at MIT.}, and Jonathan P.\ How
\thanks{Laboratory for Information and Decision Systems, Massachusetts Institute of Technology, 77 Massachusetts Ave., Cambridge, MA, USA \break
        {\tt\small \{samirw, dkkim93, shayegan, jhow\}@mit.edu}}%
}

\begin{document}

\maketitle
\thispagestyle{empty}
\pagestyle{empty}

\begin{abstract}
Multiagent reinforcement learning algorithms (MARL) have been demonstrated on complex tasks that require the coordination of a team of multiple agents to complete. Existing works have focused on sharing information between agents via centralized critics to stabilize learning or through communication to increase performance, but do not generally look at how information can be shared between agents to address the curse of dimensionality in MARL. We posit that a multiagent problem can be decomposed into a multi-task problem where each agent explores a subset of the state space instead of exploring the entire state space. This paper introduces a multiagent actor-critic algorithm and method for combining knowledge from homogeneous agents through distillation and value-matching that outperforms policy distillation alone and allows further learning in both discrete and continuous action spaces.

\end{abstract}
\section{INTRODUCTION}
Recent advances in reinforcement learning (RL) algorithms have led to novel solutions in challenging domains that range from resource management~\cite{Mao2016} to traffic control~\cite{Arel2010} to game playing~\cite{mnih15dqn,Silver2017}. The combination of RL and continuous control~\cite{Lillicrap2015a} has also found many applications in robot control. Practical robotics applications such as object manipulation or manufacturing assembly, however, often require the cooperation of multiple robots in order to efficiently complete an assigned task. Multiagent reinforcement learning (MARL) poses further challenges as the interaction of multiple agents in a shared environment leads to stability issues during the training process (a problem known as non-stationarity~\cite{Tan1993,omidshafiei2017deep}). 

A naive solution uses independent learners and apply single-agent RL algorithms for each agent. While this may prove successful in limited cases, the issue of non-stationarity invalidates many of the single-agent RL convergence guarantees~\cite{Busoniu2010}. One approach to mitigating the problem of non-stationarity combines observation and action spaces from all agents and treats the problem as a single, higher dimension, Markov decision process~\cite{Busoniu2010}. A new coordination problem, however, arises if the centralized policy needs to be executed in a decentralized manner. In a different approach, agents learn with a framework of centralized training and decentralized execution~\cite{Lowe2017, Foerster2017}. These algorithms make use of centralized value functions while maintaining independent action policies. The centralized value function stabilizes training by accounting for actions taken by other agents without requiring this extra information during execution. In general, existing works augment extra information during training to solve the issue of non-stationarity. Despite these solutions, MARL problems still suffer from the curse of dimensionality --- the state space grows exponentially with the number of agents in a given environment~\cite{Tan1993}. As each agent must independently explore a sufficient subset of the state space to learn an optimal policy, efficient exploration is crucial to minimizing training time. Current methods do not generally address this issue, leaving environments with high complexity or large numbers of agents out of reach.

\begin{figure}[t]
  \centering
  \includegraphics[width=0.9\linewidth]{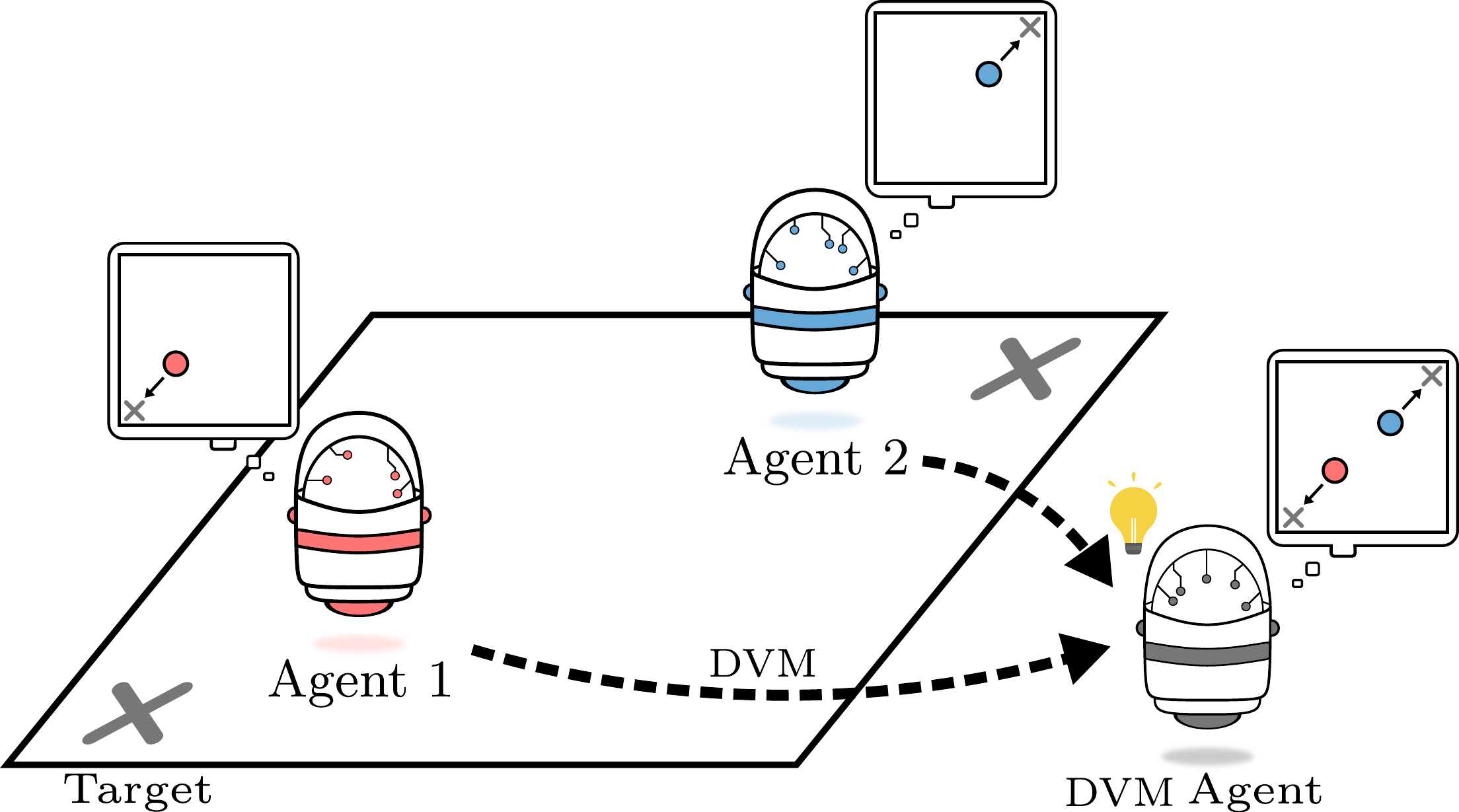}
    \caption{Overview of distillation with value matching (DVM). DVM effectively combines knowledge of Agents 1 and 2 and allows further learning.}
   \vskip -0.15in
  \label{fig:dvm-overview}
\end{figure}

In this paper, we explore how information can be shared during training to effectively combine the experiences of individual agents into new policies. We draw on the idea of policy distillation, which has previously been used to address single-agent multi-task problems~\cite{Rusu2015}. The distillation framework combines multiple policies, each individually trained on a single task, into one multi-task policy that can complete any of its parent tasks. We posit that an $n$-agent single-task problem can be re-framed as a single-agent $n$-task problem: assuming optimal behaviour from all other agents, what is the optimal policy for completing the original task as agent $i$? An example is illustrated in \cref{fig:dvm-overview} where agents must navigate to the closest target. This multiagent single-task problem can be decomposed into two symmetric single-agent tasks. In one case, an agent must learn a policy as though it were in Agent 1's position; in the other, the agent must learn a policy for Agent 2's position. The ability to share information enables agents to ``divide and conquer", where each agent only needs to learn how to complete the task from one perspective. Policy distillation, however, assumes that learning will not continue after distillation~\cite{Rusu2015}. This makes distillation alone impossible to use when the environment changes, agents are replaced, or in a scenario that would benefit from continual learning and sharing of information. 

Our work focus on teams of homogeneous agents. Even without specialized action spaces, homogeneous agents are able to learn specialized behaviours, often based on initial conditions or relative formations~\cite{Panait2005}, meaning they can be used to solve sufficiently complex MARL problems. A key property of homogeneous agents is that they are interchangeable (i.e. all agents have identical optimal policies and value functions). We exploit this property and introduce the idea of value matching --- because multiple states share identical values, we can extend the learned value in an observed state to the entire set of (possibly unobserved) states. We demonstrate that our method, distillation with value matching (DVM), is an effective method of combining agent policies and continuing to learn. We also introduce a multiagent algorithm that operates in continuous domains and can be used with policy distillation and value matching. We present empirical results that show this method converges more quickly and outperforms distillation alone in domains that vary in both complexity and the number of agents.\footnote{We plan to open-source our code upon receiving a decision on acceptance.} 
\section{BACKGROUND}

\subsection{Multiagent Reinforcement Learning}
In cooperative MARL environments, agents' actions affect the environment as well as other agents. Each agent independently observes the state and receives a reward after taking an action. Here, agents receive a joint team reward to encourage cooperation between agents. We formalize this setting as a decentralized POMDP (Dec-POMDP), which can be defined as a tuple $\langle \mathcal{I}, \mathcal{S}, \bm{\mathcal{A}}, \mathcal{T}, \bm{\Omega}, \mathcal{O}, \mathcal{R}, \gamma \rangle$~\cite{OliehoekAmato16book}.
$\mathcal{I}\!=\!\{1,\sdots,n\}$ denotes the set of agents, 
$\mathcal{S}$ denotes the set of states, 
$\bm{\mathcal{A}}\!=\!\times_{i \in \mathcal{I}} \mathcal{A}^{i}$ denotes the joint action space, 
$\mathcal{T}$ denotes the transition probability function, 
$\bm{\Omega}\!=\!\times_{i \in \mathcal{I}} \Omega^{i}$ denotes the joint observation space,
$\mathcal{O}$ denotes the observation probability function,
$\mathcal{R}$ denotes the reward function, and 
$\gamma \in [0,1)$ denotes the discount factor.\footnote{Superscript $i$ denotes a property of agent $i$. Bold symbols represent sets.}

At each time-step $t$, each agent $i$ executes an action according to its policy $a_{t}^{i}\!\sim\!\pi^{i}(o_{t}^{i};\phi^i)$ parameterized by $\phi^i$. The joint action $\bm{a}_{t}\!=\!\{a_{t}^{1},\sdots,a_{t}^{n}\}$ transitions the current state $s_{t} \in \mathcal{S}$ to the next state $s_{t+1} \in \mathcal{S}$ with probability $\mathcal{T}(s_{t+1}|s_{t},\bm{a_{t}})$. A joint observation $\bm{o}_{t+1}\!=\!\{o_{t+1}^{1}, \sdots, o_{t+1}^{n}\}$ is obtained and the team receives a shared reward $r_t\!=\!\mathcal{R}(s_{t},\bm{a}_{t})$. Actor-critic algorithms also maintain at least one critic that approximates the expected reward of a given state~\cite{sutton2018reinforcement}. An action-value critic, $Q^i(o^i_t, a^i_t; \theta^i)$, takes as input the observation and action, whereas a value critic (or value function), $V^i(o^i_t; \psi^i)$, only takes in the observation. The action-value and value critics are parameterized by $\theta^i$ and $\psi^i$, respectively. Agents' network parameters are updated during gradient steps according to the network loss functions as defined by the MARL algorithm. 

\subsection{Decentralized Actors and Centralized Critics}\label{sec:centralized-critic}
In a MARL setting, non-stationarity arises as a consequence of all agents learning simultaneously~\cite{omidshafiei2017deep}. To be specific, it is possible that
\begin{equation}
    Q^i(o^i_{t_1}, a^i_{t_1}) \neq Q^i(o^i_{t_2}, a^i_{t_2}) \suchthat t_1 \neq t_2
\end{equation}
even if $(o^i_{t_1}, a^i_{t_1}) = (o^i_{t_2}, a^i_{t_2})$ due to the fact that other agents may have taken different actions at each time-step. This has the potential to quickly destabilize learning. To address non-stationary issues in MARL, recent works~\cite{Lowe2017, Foerster2017} introduce the framework of centralized training (i.e. centralized critics) with decentralized execution (i.e. decentralized actors). Because all observations and actions are taken into account, the condition 
\begin{equation}
    Q^i(o^1_{t_1}, \mydots, o^n_{t_1}, a^1_{t_1}, \mydots, a^n_{t_1}) = Q^i(o^1_{t_2}, \mydots, o^n_{t_2}, a^1_{t_2}, \mydots, a^n_{t_2}),
\end{equation}
is satisfied at all points in time, assuming $(\bm{o_{t_1}}, \bm{a_{t_1}}) = (\bm{o_{t_2}}, \bm{a_{t_2}})$. We use MADDPG~\cite{Lowe2017}, a deterministic algorithm with a centralized critic for each decentralized actor, throughout this paper. 
Although MADDPG supports continuous action spaces, we use the discrete-action variant of MADDPG for reasons explained in \cref{sec:masac}.
\section{APPROACH}

The goal of our work is to address the need to efficiently explore a state space that grows exponentially with the number of agents. Drawing inspiration from the field of multi-task RL, we combine actor policies by creating a new, distilled policy that minimizes its Kullback-Leibler (KL) divergence with respect to all agents. We recognize that homogeneous agents learn identical policies and value functions for all permutations of agent positions within a state, and explicitly teach this to agents through value matching. Distilling policies requires that action spaces can easily be represented by a probability distribution. To apply this method in continuous spaces, we extend the Soft Actor-Critic~\cite{Haarnoja2018} algorithm to work in multiagent environments. 

\subsection{Actor Distillation} \label{sec: actor_distillation}
The intent of actor distillation is to combine the knowledge of multiple agents, regardless of the amount of overlap between what they have already learned. 

The notion of distilling neural networks was first introduced as a method for supervised model compression~\cite{Bucilua2006}. Distillation, in general, is a supervised regression process that trains a target network with an output distribution that matches the input distribution(s). Multi-task policy distillation uses the KL divergence as the loss function in training the target network~\cite{Rusu2015}. The KL divergence loss minimizes the relative entropy of one probability distribution with respect to another. We cannot minimize the KL divergence between the discrete action policies of MADDPG because they are not probability distributions. To remedy this, we apply the softmax function with temperature $\tau$ to the discrete action vectors to transform the actions into discrete categorical distributions.

Multiagent policy distillation follows the same procedure as multi-task policy distillation. Observations are sampled from agent $i$'s replay buffer and both $\pi^0(o)$ and $\pi^i(o)$ are computed (where $\pi^0$ represents a distilled policy). The softmax is applied to both outputs, and the KL loss is computed:
\begin{equation}\label{eqn:KL_loss_equation}
    L_{\text{KL}}(o, \theta^0) = \sum_{a \in \mathcal{A}}\pi^i(a\mid o)\log{\Big(\frac{\pi^i(a \mid o)}{\pi^0(a\mid o)}\Big)}.
\end{equation}
Note that $\pi$ in \cref{eqn:KL_loss_equation} does not represent the actor policy but the probability of action $a$ after taking the softmax. Once the KL loss is computed, the distilled network is optimized to minimize the loss. This is performed for each agent $i$ and repeated until the policy is considered sufficiently distilled. 

We only sample observations from the replay buffer, not actions. This allows us to query the actions from policies that have converged instead of using potentially sub-optimal actions from the replay buffer. Theoretically, agents can be distilled at any time. In practice, it is usually better to let the policies converge to a constant behaviour and then distill them. After distillation, all agents update their network parameters to match the distilled agent's parameters (they perform a hard update). This allows agents to utilize new information that might have been distilled from other agents. 

Implicit in the process of distillation is the idea of ``what" an agent knows; by sampling from the replay buffer, we only ever attempt to distill from an agent information about a state that it has visited. When agents have visited exclusively different regions of the state space (e.g., our simple spread domain in \cref{fig:spread-env}), distillation acts as a sum operation by merging information from all agents. When agents' experiences overlap and a state has been seen by multiple agents, distillation combines and minimizes variance between agents' policies. If there exists a state that no agent has seen (i.e. the environment has not been fully explored yet), the distillation process does not attempt to learn a policy for that region due to the fact that the state does not exist in the replay buffer. 

Ideally, we would like to share information more than once during the training process. If, as mentioned, agents' policies before distillation do not contain sufficient information to distill an optimal policy, we would like to gather more information and distill once more. The environment (e.g. obstacles, objectives) may change, and agents must then learn slightly different policies. Unfortunately, without changing the value function, the mismatch between distilled actor and original critic in previously unobserved states leads to divergence, and a catastrophic ``forgetting" of the information that was distilled may occur. To continue learning after distillation, it is necessary to update the critic such that it reflects the new actor policy, which we explain how to do in the next section.

\subsection{Homogeneous Value Matching}

Agents which are homogeneous in their observation and action spaces have the characteristic of being interchangeable within the same environment. Without explicit specialization as a result of differing capabilities, the optimal policy for all agents with a given observation is the same~\cite{Panait2005}. If the reward function is the same for each agent (equivalent to a joint reward function), then it follows that the Q-function and value functions will also be the same. We take advantage of these properties to minimize the amount of exploration necessary through homogeneous value matching. 

A centralized critic receives information regarding observations and actions from all agents. More importantly, the order of the input to the value function is often fixed and of the form $V^i(o^1, \ldots, o^n)$ for $n$ agents~\cite{Lowe2017}. Consider a two agent environment where the agents have learned the value of state A. Let state B represent a new state where the agents have swapped positions with one another. The values of the two states can be represented with $V^i(o^1_A, o^2_A)$ and $V^i(o^1_B, o^2_B)$, respectively. Due to homogeneity, we know that the values of states A and B should be equal. Furthermore, if agent 1's new observation is the same as agent 2's original observation (and vice versa), the value functions become
\begin{equation}
    V^i(o^1_A, o^2_A) = V^i(o^1_B, o^2_B) = V^i(o^2_A, o^1_A).
\end{equation}

Enforcing agent observations to be identical from swapped positions (with all else constant) is a reasonable and practical design choice. Common sensors used in robotic applications such as radar and lidar do not differentiate between and identify specific objects, but rather return simple information relative to the user --- both of these sensors return equal observations if two (or more) objects swap positions. In a simulation environment built for testing MARL algorithms~\cite{lowe2017multi}, observations are not purely relative and do not meet this requirement. We modified the environment to return the positions of other agents as sorted by their relative headings, emulating how a lidar would return data.

When relative observations meet this criterion, the following property holds:
\begin{equation}
    V^i(o^1, \ldots, o^n) = V^i(X) \quad \forall X \in \mathfrak{S}(o^1, \ldots, o^n),
\end{equation}
where $\mathfrak{S}$ represents the symmetric group of the joint observation space containing all permutations of the individual observations.

Normally, a set of agents would have to experience both $(o^1, o^2)$ and $(o^2, o^1)$ during the training process in order to approximate the value or expected return of both states. We introduce the idea of value matching: once the value of one state is learned, train the critic to assign the same value for all states in the symmetric group of the observation. This is done by using supervised learning to train a new ``value-matched" critic network. The process of distillation and value matching is referred to as DVM. 

To train a new value network, we sample an observation from the replay buffer and query the value network of an agent, $V^i(\bm{o})$ and the distilled value network, $V^0(\bm{o})$. We use a mean-squared-error-loss (MSE) to train the distilled network,
\begin{equation} \label{eq:mse_loss}
    L_{\text{MSE}}(\bm{o}, \psi^0) = \sum_{X \in \mathfrak{S}(\bm{o})} \left\lVert V^i(\bm{o}) - V^0(X) \right\rVert^2_2,
\end{equation} where $\psi$ represents the parameters of the matched value function. Similar to a distilled actor, this ``value matched" value function can now be used by the agent to represent its knowledge of the state space, even if it did not explicitly visit many regions of it.

MADDPG and many actor-critic algorithms do not parameterize and make use of a value function. Instead, they use an action-value function, or a Q-function that takes as input both an observation and action. As denoted in \cref{sec:centralized-critic}, centralized critics have access to observations and actions taken by \textit{all} agents. The Q-function then takes the form $Q(\bm{o}, \bm{a})$. The idea of value matching also works on Q-functions and is performed in a similar manner. Both observations and actions are sampled from the replay buffer and used to calculate the MSE loss. It is critical for action-value functions that a given $(o, a)$ is treated as a pair; permutations of $\bm{o}$ and $\bm{a}$ must be consistent with each other to ensure that the observations match the actions they resulted in.

Despite this process occurring with policy distillation, it is incorrect to refer to value matching as ``distilling value functions". Distillation attempts to combine information from multiple policies. Even though MADDPG has a centralized critic for each agent, they all have access to the same information. Each agent can thus perform value-matching with it's own critic. As a result, value matching can be applied to agents with heterogeneous capabilities or differing reward functions. Because we have homogeneous agents (same critic input size) and a joint reward function (same reward scale), we choose to create one value-matched critic that all agents will use. This also follows from the assumption that a team of homogeneous agents will eventually converge on the same optimal value function~\cite{Panait2005}. 

\subsection{Multiagent Soft Actor-Critic} \label{sec:masac}

\subsubsection{Background}

We first review key concepts in Soft Actor-Critic (SAC)~\cite{Haarnoja2018} that are related to developing our new method. SAC is an off-policy actor-critic algorithm built on a maximum entropy framework to balance stability and exploration. Maximum entropy RL algorithms attempt to maximize both the expected reward and the expected entropy of the learned policy~\cite{Haarnoja2017}. 
Fundamental to SAC is the \textit{stochastic} actor. In SAC, the actor policy outputs a probability distribution from which actions can be sampled during exploration. SAC can also be used to learn in continuous action spaces.

SAC utilizes a soft value function $V(o_t;\psi)$, a soft Q-function $Q(o_t, a_t;\theta)$, and a policy $\pi(a_t|s_t;\phi)$. 
During training, the soft value function is trained to minimize the loss function
\begin{equation} \label{eq:single_agent_v_loss}
\begin{split}
J_V(\psi) = \mathbb{E}_{o_t \sim \mathcal{D}}\big[ \tfrac{1}{2}\big( V(o_t;\psi) - \mathbb{E}_{a_t \sim \pi}[Q(o_t, a_t;\theta) \\
- \log \pi(a_t|o_t;\phi)]\big)^2\big]
\end{split}
\end{equation}
where $\mathcal{D}$ represents the replay buffer of previously experienced observations and actions~\cite{Lin1992}. 
The soft Q-function is similarly trained by minimizing the soft Bellman residual
\begin{equation} \label{eq:single_agent_q_loss}
J_Q(\theta)\! =\! \mathbb{E}_{o_t, a_t, r_t, o_{t+1} \sim \mathcal{D}}\big[ \tfrac{1}{2}\big( Q(o_t, a_t;\theta) \!-\!\hat{Q}(o_t, a_t)\big)^2 \big]
\end{equation}
\begin{equation} \label{eq:q_hat}
    \hat{Q}(o_t, a_t) = r_t + \gamma V(o_{t+1};\bar{\psi})
\end{equation}
where $\bar{\psi}$ represents a target value network.

\subsubsection{Algorithm}

We introduce Multiagent Soft Actor-Critic (MA-SAC), an extension of SAC that can be utilized in multiagent environments. MA-SAC makes use of a centralized critic and a centralized value function to mitigate the problem of non-stationarity in MARL (\cref{sec:centralized-critic}). This is done by augmenting the input of the Q-function and value function with the observations and actions of all other agents. Using the set of all observations and actions instead, the soft value function and Q-function loss functions for a specific agent $i$ become
\begin{equation} \label{eq:multi_agent_v_loss}
\begin{split}
J_{V^i}(\psi^i) = \mathbb{E}_{\bm{o_t} \sim \mathcal{D}}\big[ \tfrac{1}{2}\big( V^i(\bm{o_t};\psi^i) - \mathbb{E}_{\bm{a_t} \sim \pi^i}[Q^i(\bm{o_t}, \bm{a_t};\theta^i) \\
- \log \pi^i(\bm{a^i_t} \mid \bm{o^i_t};\phi^i)]\big)^2\big],
\end{split}
\end{equation}
and
\begin{equation} \label{eq:multi_agent_q_loss}
J_{Q^i}(\theta^i) = \mathbb{E}_{\bm{o_t}, \bm{a_t}, r_t, \bm{o_{t+1}} \sim \mathcal{D}}\big[ \tfrac{1}{2}\big( Q^i(\bm{o_t}, \bm{a_t}; \theta^i) - \hat{Q}^i(\bm{o_t}, \bm{a_t})\big)^2 \big]
\end{equation}
\begin{equation} \label{eq:q_hat_multi}
    \hat{Q}^i(\bm{o_t}, \bm{a_t}) = r_t + \gamma V^i(\bm{o_{t+1}};\bar{\psi^i})
\end{equation}

The distillation process explained in \cref{sec: actor_distillation} relies on manipulating probability distributions. For actors with discrete action spaces, any general purpose MARL framework (e.g. MADDPG, COMA) can be used with our algorithm by turning the action space into a discrete categorical distribution. The continuous variant of MADDPG, however, returns a single deterministic action for a given observation. Because extracting the continuous distribution over the action space is generally intractable (as is computing the KL divergence between two arbitrarily continuous distributions), we look to MA-SAC as policies can be inherently represented with probability distributions. We implement policies that output a Gaussian with mean and variance to represent a continuous action probability distribution. An overview of the entire algorithm, including DVM, can be seen in \cref{alg:pseudo}.

\begin{algorithm}[t]
	\caption{Multiagent Soft Actor-Critic with DVM}\label{alg:pseudo}  
	\begin{algorithmic}[1]
	    \State Initialize parameters $\bm{\psi}$, $\bm{\bar{\psi}}$, $\bm{\theta}$, $\bm{\phi}$. 
	    \State Initialize distilled parameters $\psi^0$, $\bar{\psi^0}$, $\theta^0$, $\phi^0$. 
	    \State Initialize replay buffer $\mathcal{D}$
		\For{each training iteration}
		    \For{each environment step $t$}
		        \State Step in environment with $\bm{a_t} \sim \bm{\pi}$
		        \State Store experience in replay buffer $\mathcal{D}$
            \EndFor
            \For{each gradient step}    
                \For{each agent $i$}
                    \State Update network $\phi^i$ according to SAC~\cite{Haarnoja2018}
                    \State Update networks $\psi^i, \hat{\psi}^i$ according to \cref{eq:multi_agent_v_loss}
                    \State Update network $\theta^i$ according to \cref{eq:multi_agent_q_loss}
        		\EndFor
		    \EndFor
		\EndFor
		\For{$j$ it\tikzmark{left}erations}
		    \For{each agent $i$}
		        \State\tikzmark{top}Distill policies ($\pi^0$, $\pi^i$) according to \cref{eqn:KL_loss_equation}
		        \For{each permutation of agents}
                    \State Value-match ($Q^0$, $Q^i$) according to \cref{eq:mse_loss}
                \EndFor\tikzmark{bottom}
		    \EndFor
		\EndFor
		\vskip -.2in
	\end{algorithmic}
	\AddNote{top}{bottom}{left}{DVM}
\end{algorithm}

\section{RELATED WORK}
The fundamental issue we address is how to effectively transfer information between agents within and between tasks. \cite{Taylor2013} provides a thorough survey on transfer learning RL, but recent work in other fields address the same issue. Imitation learning, for example, has been studied to learn from demonstrations~\cite{argall2009survey, ross2011reduction}, and inverse learning to extract the hidden reward function~\cite{ng2000algorithms, abbeel2004apprenticeship}. The work in these fields, however, focus primarily on single-agent problems.

Past literature on transferring information in MARL is also generally limited to transferring information between tasks rather than between agents within a single task. \cite{le2017coordinated} attempts to perform imitation learning in a multiagent environment. \cite{boutsioukis2011transfer} proposes a method to use the solution of one learned MARL problem as an initialization of another. \cite{Taylor2013} uses transfer learning between agents in MARL to accelerate convergence to an optimal solution, but what information to share must be selected by hand. Although \cite{omidshafiei2017deep} uses policy distillation for multi-task multiagent problems, no information is shared between agents to improve learning performance. 
Work in the field of agent advising attempt to answer the question of how to incorporate outside knowledge during the learning process. The core concept is to have an advisor provide guidance to an agent during the learning process via instruction or advice~\cite{wiewiora2003principled, clouse1992teaching}. This idea has been applied to MARL as well, where agents can advise each other. \cite{da2017simultaneously} demonstrates how agents in MARL can learn and advise simultaneously, and \cite{omidshafiei2018learning} extends this idea by having agents learn policies about the correct way to provide advice. 

An important distinction should be made between distillation and the utilization of centralized value functions. While the addition of extra information in centralized value functions can be thought of as agents sharing information, the information shared does not inform how a recipient's policy should change. Similarly, agents can cooperate by learning to communicate~\cite{Lowe2017, Foerster2017, sukhbaatar2016learning, mordatch2017emergence, LazaridouPB16b}, but they do not communicate with the express purpose of improving other agents' policies.
\begin{figure}[t]
  \centering
  \includegraphics[width=0.43\textwidth]{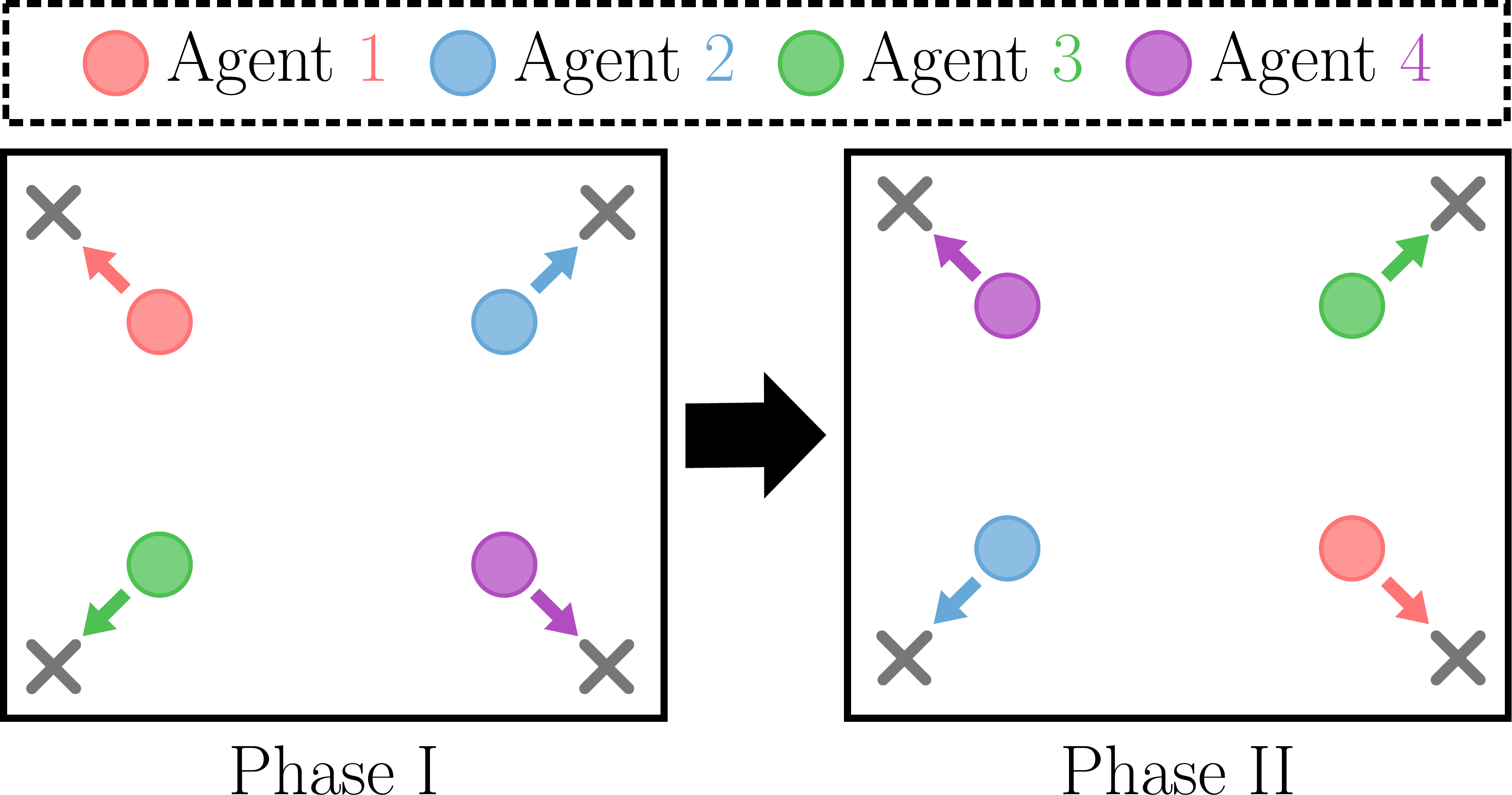}
  \caption{Spread domain example with four agents. The goal of the agents is to cover all targets as quickly as possible. Agents are initialized in specific configurations depending on the phase. Learning begins in Phase I, then DVM is applied and learning continues in Phase II.}
  \label{fig:spread-env}
  \includegraphics[width=0.49\textwidth]{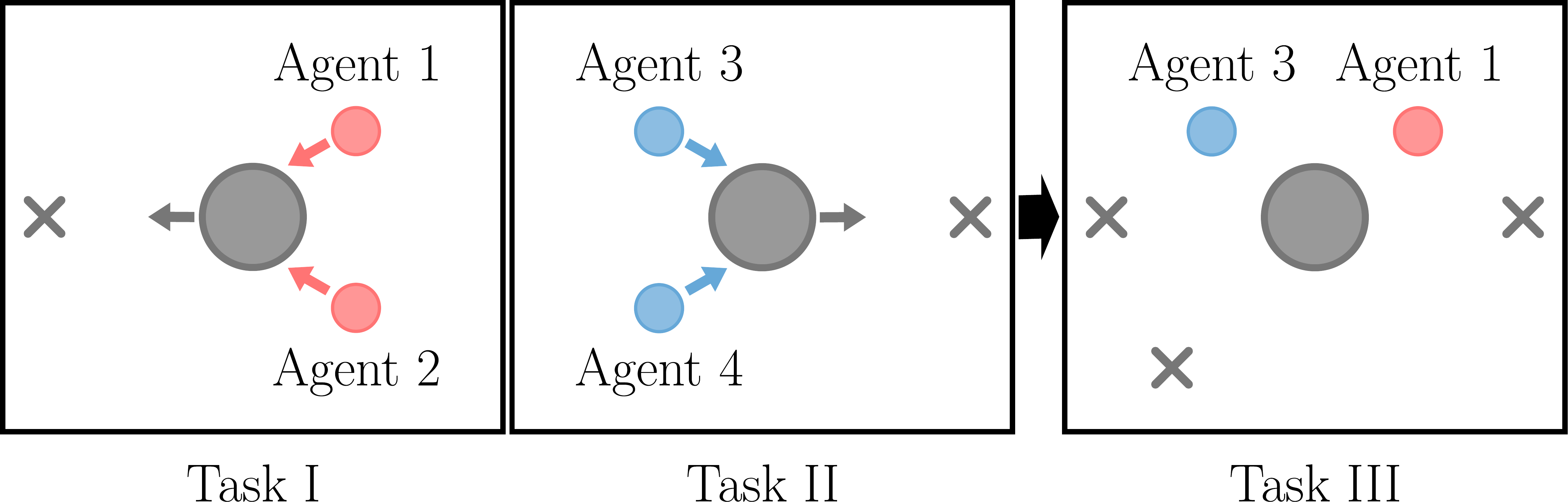} 
  \caption{Cooperative push box domain. Red and blue circles represent agents. The large grey circle represents a box that requires both agents applying force to move. The goal is to reach the target as quickly as possible. Agents are pre-trained in Tasks I and II where they learn to move the box to a specific target. DVM is applied, and agents learn in Task III. The target location in Task III is randomly selected from the locations in the original tasks, as well as a new location.}
  \label{fig:two-box}
\end{figure}

\section{RESULTS}
We evaluate the performance of DVM in a number of single-task domains that vary in complexity and number of agents (the two-agent, three-agent, and four-agent spread domains) as well as a challenging multi-task domain (push box domain).

\subsection{Domains}

The domains are built on top of OpenAI's multiagent particle environments~\cite{lowe2017multi}. Environments support both discrete and continuous observation/action spaces, and allow for easy modification of existing environments. 

\begin{figure*}[t]
\captionsetup[subfigure]{skip=-2pt} 
  \begin{subfigure}[b]{0.33\textwidth}
     \centering
     \includegraphics[trim=10 0 35 23,clip,width=\textwidth]{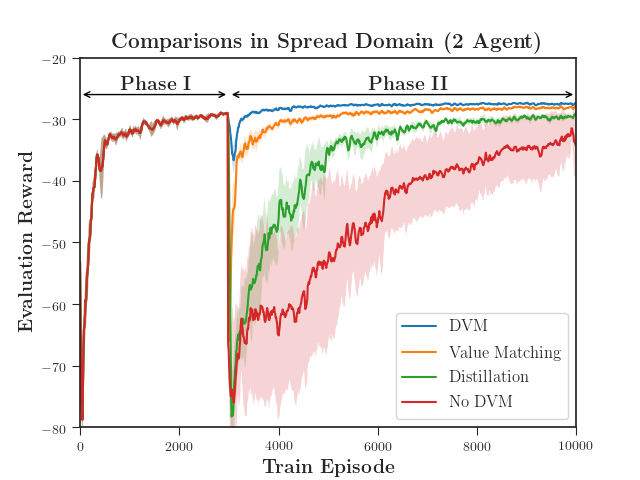}
  \caption{}
  \label{fig:spread-two-discrete-result}
  \end{subfigure}
  \begin{subfigure}[b]{0.33\textwidth}
     \centering
     \includegraphics[trim=10 0 35 23,clip,width=\textwidth]{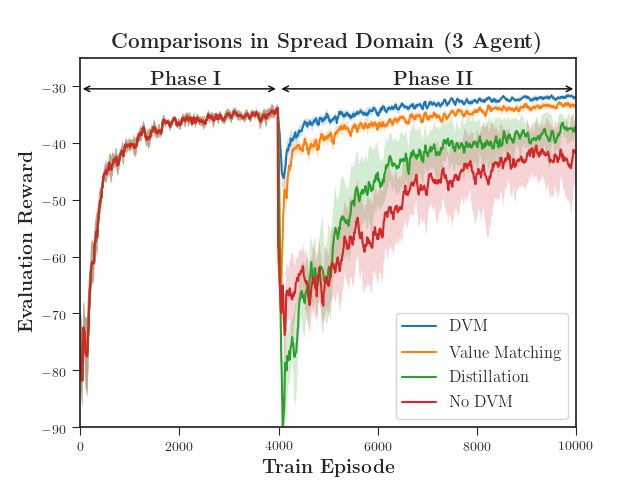}
  \caption{}
  \label{fig:spread-three-discrete-result}
  \end{subfigure}
  \begin{subfigure}[b]{0.33\textwidth}
     \centering
     \includegraphics[trim=10 0 35 23,clip,width=\textwidth]{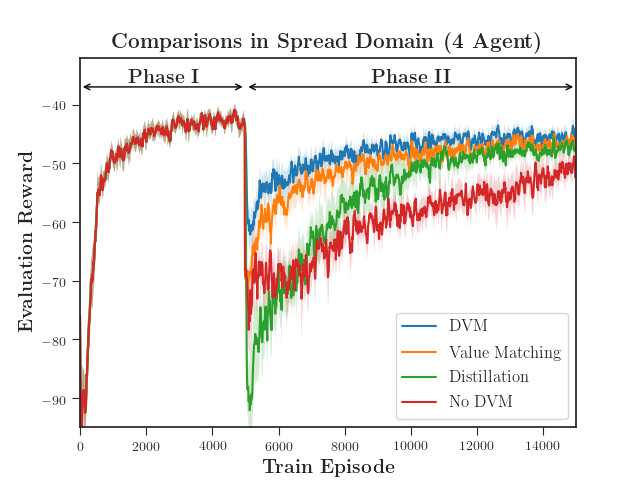}
  \caption{}
  \label{fig:spread-four-discrete-result}
  \end{subfigure}
  \caption{DVM Performance in the Discrete Spread Domain \textbf{(a)} Two-Agent.
  \textbf{(b)} Three-Agent. \textbf{(c)} Four-Agent.}
	\label{fig:overview_fig_discrete}
\end{figure*}

\begin{figure*}[t]
\captionsetup[subfigure]{skip=-2pt} 
  \begin{subfigure}[b]{0.33\textwidth}
     \centering
     \includegraphics[trim=10 0 35 23,clip,width=\textwidth]{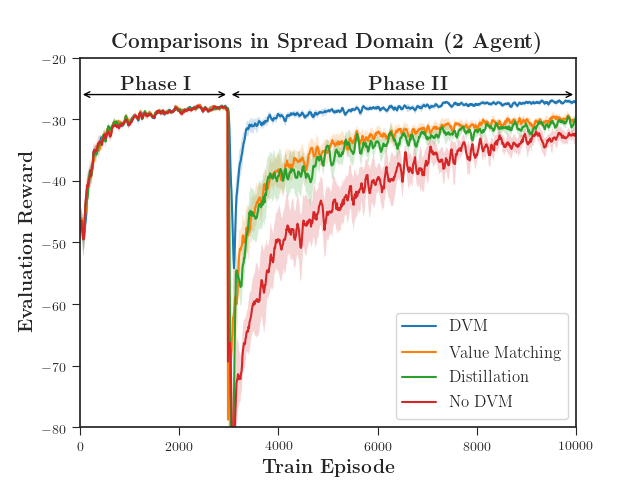}
  \caption{}
  \label{fig:spread-two-continuous-result}
  \end{subfigure}
  \begin{subfigure}[b]{0.33\textwidth}
     \centering
     \includegraphics[trim=10 0 35 23,clip,width=\textwidth]{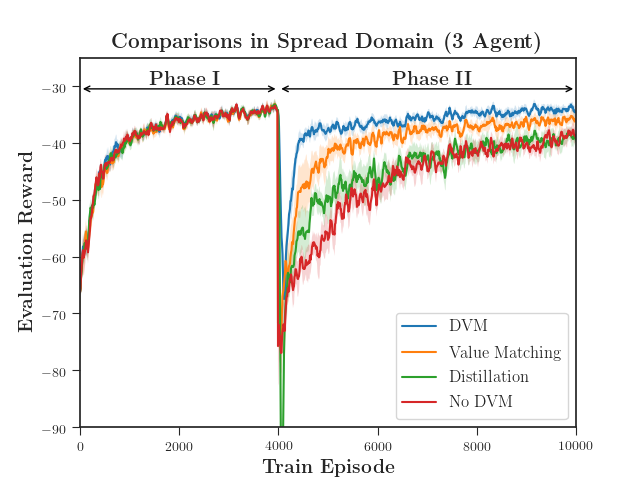}
  \caption{}
  \label{fig:spread-three-continuous-result}
  \end{subfigure}
  \begin{subfigure}[b]{0.33\textwidth}
     \centering
     \includegraphics[trim=10 0 35 23,clip,width=\textwidth]{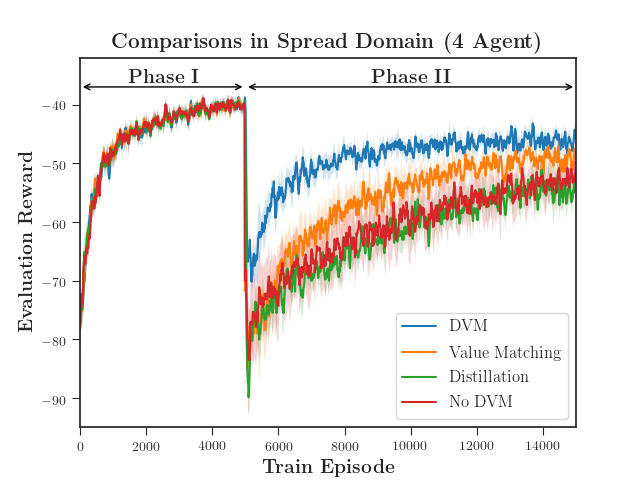}
  \caption{}
  \label{fig:spread-four-continuous-result}
  \end{subfigure}
  \caption{DVM Performance in the Continuous Spread Domain \textbf{(a)} Two-Agent.
  \textbf{(b)} Three-Agent. \textbf{(c)} Four-Agent.}
	\label{fig:overview_fig_continuous}
\end{figure*}

\subsubsection{Spread Domain}
$n$ agents and $n$ targets comprise the spread domain. The four-agent spread domain is shown in \cref{fig:spread-env}. Agents are initialized randomly within the environment and receive a joint reward based on how far any agent is from each target. The goal is to cover all targets as quickly as possible. The two-agent and three-agent spread domains are similar but with fewer agents and targets. 

The initial position of the agents is restricted to the four quadrants of the environment with one agent in each quadrant. Although agents are free to move throughout the environment, the starting location artificially restricts agents to learn to cover the target in the quadrant they start in. 

The quadrant in which an agent $i$ is initialized depends on the Phase of the experiment. Agents are initially trained in Phase I with the fixed initial quadrants shown. DVM is performed, and agents then continue training in Phase II. In the two-agent and three-agent domains, Phase II also has the fixed initial quadrants as shown. The four-agent domain in Phase II is slightly different in that the initial quadrants are not fixed, but randomized. This introduces more complexity as agents originally trained with one configuration must now learn a policy for all twenty-four possible start configurations. 

\subsubsection{Push Box}
A heavy box and two agents comprise the push box domain. Two agents must work together to manipulate a box around the environment in the push box domain (\cref{fig:two-box}). A box is initialized at a fixed starting point in the center of the environment, and the agents are initialized in the environment randomly around the box. Agents receive a joint reward $r_{t}=-||\text{loc(target)} - \text{loc(box)}||^2_2$. The goal is to move the box to the target as quickly as possible.

This domain is significantly more difficult as the box can be moved \textit{if and only if} the two agents apply a force to (move) the box together. This ensures that the goal cannot be completed without cooperation. Additionally, agents must coordinate to apply forces such that the box moves in a straight line to the target and maximizes reward. This domain presents an issue of delayed reward as the reward does not change until the box is acted upon by both agents.

We test this domain in a more traditional multi-task manner. Agents 1 and 2 are trained on Task I and learn to push the box to a target on the left. Agents 3 and 4 are trained on Task II and push the box to the right. DVM is applied to Agents 1 and 3 --- one agent from each task --- and tested in Task III. In Task III, the location of the target is chosen at random from the first two tasks and an additional location equally far away from the box. 

This multi-task domain allows for a direct comparison with multi-task policy distillation and demonstrates the advantage of continuing to learn when the environment changes (Task III contains a target not seen in Tasks I and II). 

\subsection{Simulation Results}

We train on neural networks with two hidden layers and 256 units per layer. We use epsilon-greedy exploration noise for MADDPG and an $\alpha = 0.1$ for MA-SAC. The ADAM optimizer is used to update the networks in both algorithms with a learning rate of 0.01 and 0.0003, respectively. The batch size is 1024 for both algorithms. The policy is evaluated every 10 steps, either with no noise for MADDPG or deterministically for MA-SAC.

We ran 10 seeds for each experiment. The dark lines represent the mean values, and the shading represents one-half standard deviation from the mean. We compared DVM (with 2048 iterations) against three baselines to evaluate its performance.

\begin{itemize}
    \item \textbf{No DVM} - no distillation or value matching is applied between phases I and II. 
    \item \textbf{Distillation} - regular policy distillation (2048 iterations) is applied with no value matching.
    \item \textbf{Value Matching} - value matching (2048 iterations) is applied to the critic with no distillation.
\end{itemize}

\subsubsection{Spread Domain}
We evaluated the performance of DVM with both a discrete action space using MADDPG (\cref{fig:overview_fig_discrete}) and a continuous action space using MA-SAC (\cref{fig:overview_fig_continuous}). DVM outperforms all baselines, converging to a better policy more quickly than the rest. Distillation alone fails to maintain the distilled policy as the critic causes the policy to diverge. As the complexity increases, distillation without value matching does only marginally better than not distilling at all (\cref{fig:spread-four-continuous-result}). Interestingly, value matching alone performs quite well relative to DVM. We hypothesize that, if the continuation of learning is necessary, updating the critic is far more important the distilling an actor.

\begin{figure}[t]
  \centering
  \includegraphics[width=0.45\textwidth]{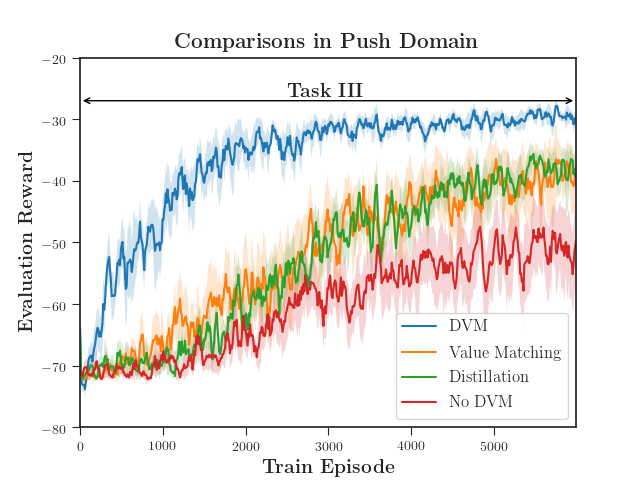}
  \caption{DVM Performance in the Push Box Domain. DVM outperforms all baselines, converging to the optimal policy even though Task III includes a target location not present in either of the pre-training tasks.} 
  \label{fig:push}
\end{figure}

\begin{figure}[t]
  \centering
  \includegraphics[width=0.45\textwidth]{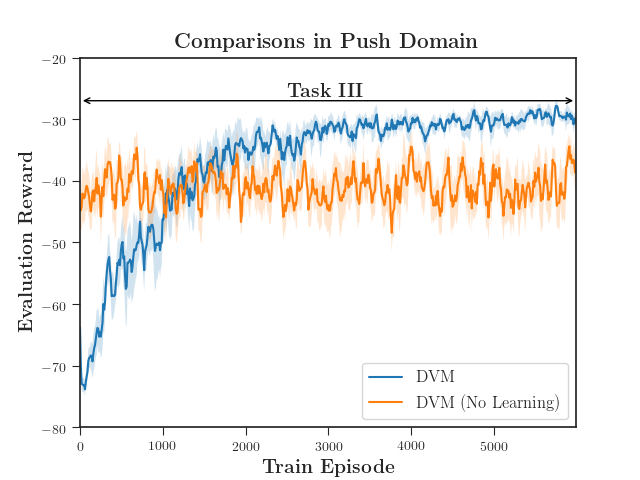}
  \caption{DVM Performance with and without learning. Without learning, the policy does not diverge but cannot learn the optimal policy or how to reach the new target location. DVM starts worse, but quickly reaches and outperforms the baseline.} 
  \label{fig:push_eval}
\end{figure}

\subsubsection{Push Box}
We evaluate the push box domain with a discrete action space with MADDPG (\cref{fig:push}). As before, DVM outperforms all baselines. Both distillation-only and value-matching-only methods perform similarly, but do not reach an optimal policy within the number of testing periods. We speculate that the high variance in the three baselines is a result of adding the third target in Task III. 

To evaluate the effect of a changing environment, we compare DVM with and without learning in Task III (\cref{fig:push_eval}). When the agents do not update their policies, there is no possibility of the policies diverging. This is equivalent to multi-task policy distillation and simply evaluating the performance of the distilled policy. 

Though the agents which continue learning drop in evaluation score as the policies and critics reconcile their updates, they quickly recover and converge on an optimal policy. Even though the environment has changed with the introduction of the third target, the DVM agents learn the correct policy and return a steady-state reward higher than the baseline. 

\begin{figure}[t]
  \centering
  \includegraphics[width=0.50\textwidth]{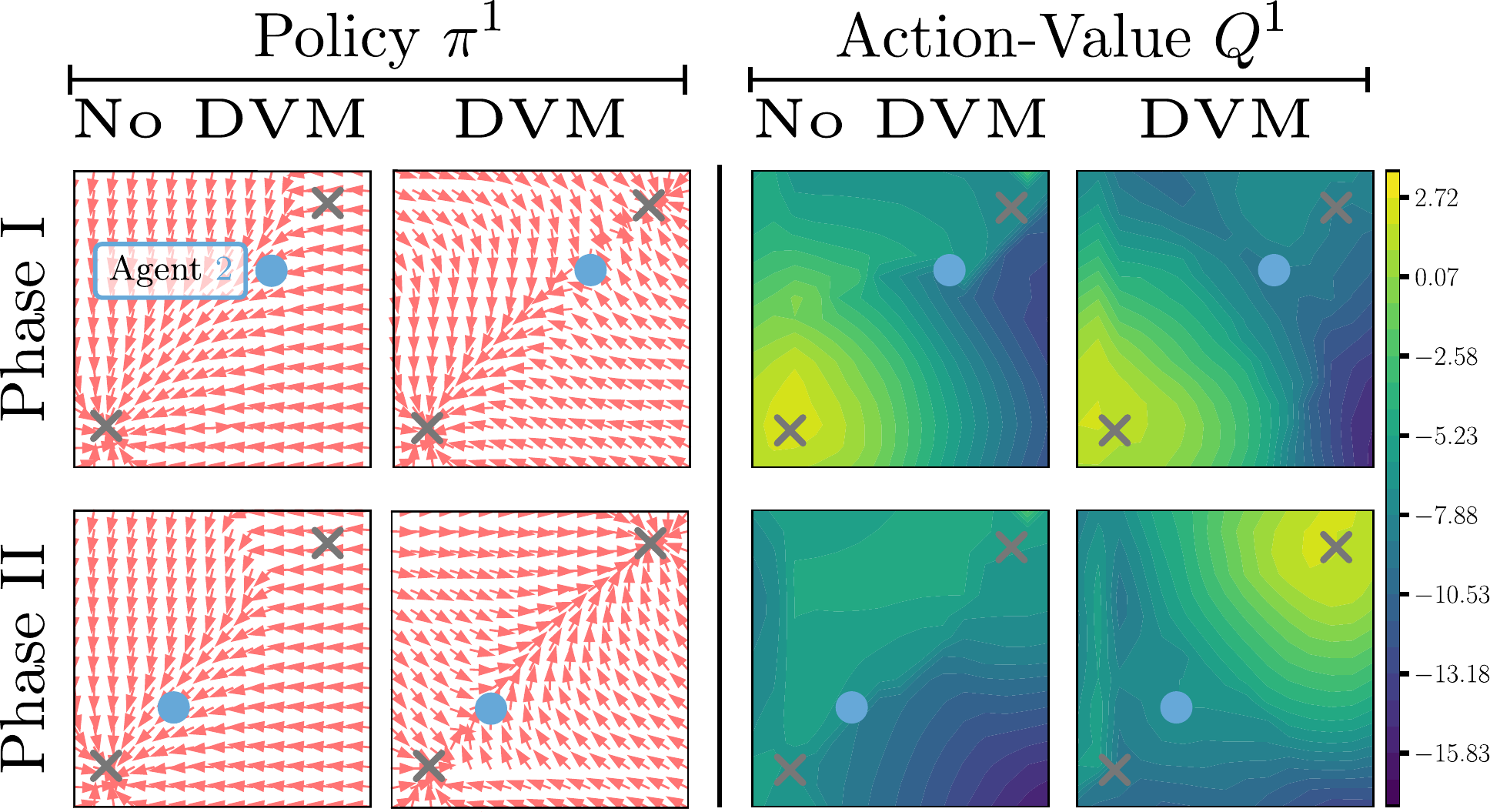}
  \caption{Analysis of Agent 1's policy and action-value function before and after DVM. Because Agent 1 was trained in Phase I, the policy and action-value function are correct in Phase I configuration. In the Phase II configuration, the action-value function is uninformed while the policy is incorrect. After DVM, the information from Phase I is retained and the correct policy and action-value functions for Phase II are learned even though the agent has never been in a Phase II configuration.} 
  \label{fig:analysis}
\end{figure}

\subsection{Analysis}

We analyzed the policies and action-value functions of agents in the two-agent spread domain before and after DVM (\cref{fig:analysis}). Agent 1 (red agent) starts in the lower left (LL) corner in Phase I, and Agent 2 (blue agent) starts in the upper right (UR) corner. In Phase II, the starting configuration is flipped. The plots show the policies and critic values for Agent 1 in the configurations seen in both phases, as well as Agent 2's position.

As the agents train in Phase I, Agent 1 learns to navigate to the LL target. Without the distillation aspect of DVM, Agent 1 continues to navigate to the LL target even though it is much closer to the UR (Phase II, No DVM Policy). Distilling the policies resolves this. After DVM, Agent 1 maintains the correct policy from Phase I while also learning to navigate to the UR target in Phase II (Phase II, DVM policy).

The action-value plot represents the effect of value matching. Agents will generally follow a policy that climbs the gradient. As before, the gradient of the action-value function is correct for the phase it was trained in. In the Phase II configuration, the action-value function is much less defined, and definitively does not lead the agent to the UR target. After value matching, the gradients in both configurations lead the agent to the correct targets (Phase II, DVM action-value).
\section{CONCLUSION}
We present DVM, a method for combining knowledge between a team of homogeneous agents. We leverage the properties of homogeneous agents to introduce the idea of value matching, allowing us to inform an agent's critic of states the agent has not yet visited. Combined with policy distillation, we show that DVM enables agents to visit different regions of the state space, combine the information, and continue learning. In order to perform DVM in continuous action spaces, we introduce MA-SAC, a multiagent variant of the SAC algorithm. Empirical results demonstrate that DVM outperforms distillation or value matching alone and that DVM enables agents to learn in potentially changing environments. Future work in this area include extending the current framework to support teams of heterogeneous agents and developing a paradigm to continually share information during learning.

\section*{Acknowledgement}

This work was supported by Boeing Research \& Technology, IBM (as part of the MIT-IBM Watson AI Lab initiative), and AWS Machine Learning Research Awards program. 

\balance

\bibliographystyle{IEEEtran} 
\bibliography{references}  

\addtolength{\textheight}{-12cm}   



\end{document}